\documentclass[letterpaper]{article} 
\usepackage{aaai24}  
\usepackage{times}  
\usepackage{helvet}  
\usepackage{courier}  
\usepackage[hyphens]{url}  
\usepackage{graphicx} 
\usepackage{natbib}  
\usepackage{caption} 
\usepackage{algorithm}
\usepackage{algorithmic}
\usepackage{booktabs}
\usepackage{multirow}
\usepackage{arydshln}
\usepackage{makecell}
\usepackage{xcolor}
\usepackage[tight]{subfigure}
\usepackage{enumerate}
\usepackage{microtype}
\usepackage[shortcuts]{extdash}

\definecolor{first_one}{RGB}{207,216,197}
\definecolor{second_one}{RGB}{214,198,144}
\definecolor{third_one}{RGB}{189,197,165}
\definecolor{fourth_one}{RGB}{117,131,100}
\definecolor{fifth_one}{RGB}{71,82,47}
\definecolor{red_one}{RGB}{230,169,180}
\usepackage{newfloat}
\usepackage{listings}
\DeclareCaptionStyle{ruled}{labelfont=normalfont,labelsep=colon,strut=off} 
\floatstyle{ruled}
\newfloat{listing}{tb}{lst}{}
\floatname{listing}{Listing}
\title{Mitigating Large Language Model Hallucinations via Autonomous \\ Knowledge Graph-based Retrofitting}
\author {
    Xinyan Guan\textsuperscript{\rm 1,\rm 3}\equalcontrib,
    Yanjiang Liu\textsuperscript{\rm 1,\rm 3}\equalcontrib,
    Hongyu Lin\textsuperscript{\rm 1},
    Yaojie Lu\textsuperscript{\rm 1},\\
    Ben He\textsuperscript{\rm 1,\rm 3},
    Xianpei Han\textsuperscript{\rm 1,\rm 2},
    Le Sun\textsuperscript{\rm 1,\rm 2}
}
\affiliations {
    \textsuperscript{\rm 1}Chinese Information Processing Laboratory
    \textsuperscript{\rm 2}State Key Laboratory of Computer Science\\ Institute of Software, Chinese Academy of Sciences, Beijing, China\\
    \textsuperscript{\rm 3}University of Chinese Academy of Sciences, Beijing, China
\\
    \{guanxinyan2022,hongyu,luyaojie,xianpei,sunle\}@iscas.ac.cn, \\
    liuyanjiang22@mails.ucas.ac.cn,benhe@ucas.edu.cn
}

\usepackage{bibentry}

\begin{document}

\maketitle

\begin{abstract}
Incorporating factual knowledge in knowledge graph is regarded as a promising approach for mitigating the hallucination of large language models (LLMs). 
Existing methods usually only use the user's input to query the knowledge graph, thus failing to address the factual hallucination generated by LLMs during its reasoning process. To address this problem, this paper proposes Knowledge Graph-based Retrofitting (KGR), a new framework that incorporates LLMs with KGs to mitigate factual hallucination during the reasoning process by retrofitting the initial draft responses of LLMs based on the factual knowledge stored in KGs. Specifically, KGR leverages LLMs to extract, select, validate, and retrofit factual statements within the model-generated responses, which enables an autonomous knowledge verifying and refining procedure without any additional manual efforts. 
Experiments show that KGR can significantly improve the performance of LLMs on factual QA benchmarks especially when involving complex reasoning processes, which demonstrates the necessity and effectiveness of KGR in mitigating hallucination and enhancing the reliability of LLMs.

\end{abstract}

\section{Introduction}
\begin{figure*}[t]
\centering
\includegraphics[width=1.0\textwidth]{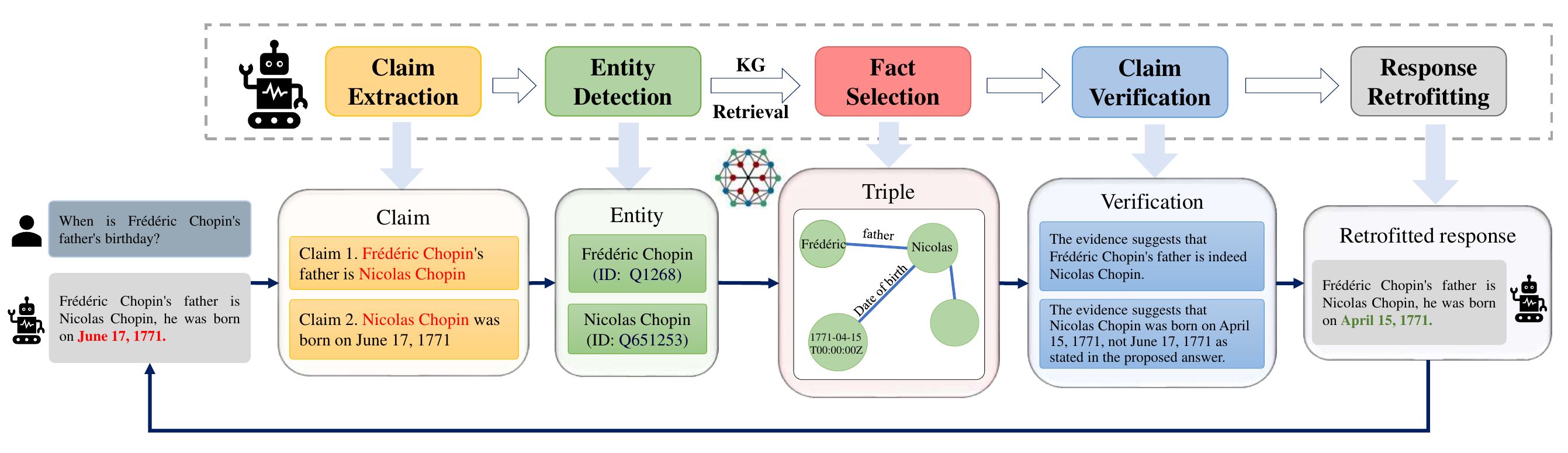}
\caption{An overview of KGR, our framework consists of five components: (1) claim extraction, (2) entity detection and KG retrieval, (3) fact selection, (4) claim verification, (5) response retrofitting. The core component of these five steps remains the LLM. Given a question and draft response as input, our framework can iteratively mitigate factual errors in LLM's response.}
\label{fig:flow}
\end{figure*}

Large Language Models (LLMs) have gained increasing prominence in artificial intelligence. The emergence of potent models such as ChatGPT~\cite{2022chatgpt} and LLaMA~\cite{touvron2023llama} has led to substantial influences on many areas like society, commerce, and research. However, LLMs still suffer from severe \textit{factual hallucination} problems, i.e., LLMs can frequently generate unsupported false statements regarding factual information due to their lack of intrinsic knowledge~\cite{ji2023survey}. For example, in Figure~\ref{fig:flow}, ChatGPT fails to provide an accurate response to the query ``\textit{When is Frédéric Chopin's father's birthday?}'' due to a wrong belief that Nicolas Chopin's birthday is on June 17, 1771. Factual hallucination poses a severe challenge for LLM applications, particularly in real-world situations where factual accuracy holds significance. Consequently, the endeavor to alleviate factual hallucinations in LLMs has become a research hotspot in NLP field~\cite{liu2021faithfulness,kang2020improved}.

On the other hand, Knowledge Graphs (KGs) store a substantial amount of high-quality factual information, which can significantly alleviate factual hallucination if incorporated with LLMs.
For example, in Figure~\ref{fig:flow}, we can retrofit the erroneous statement ``\textit{Nicolas Chopin was born on June 17, 1771}'' by referring to the provided factual knowledge ``\textit{(Nicolas Chopin, date of birth, 1771-04-15T00:00:0)}'' in Wikidata. 
Recent work has focused on integrating LLMs with KGs by retrieving the entities in the query within knowledge graphs. Then the obtained factual triples are utilized as an additional context for LLMs to enhance their factual knowledge~\cite{baek2023knowledge, Chase_LangChain_2022}. Unfortunately, these approaches are limited to retrieving factual knowledge relevant to entities explicitly mentioned within the given query. However, the fundamental capability of large language models involves intricate and multi-step reasoning. Such reasoning processes often necessitate the validation and augmentation of factual knowledge that may be employed during the reasoning process. For example, in the case shown in Figure~\ref{fig:flow}, LLM fails to answer the question because it requires an intermediate knowledge about ``\textit{Nicolas Chopin was born on April 15, 1771}''. However, such information does not refer to entities appearing in the query. As a result, previous approaches are inadequate in addressing the factual hallucination appearing in the reasoning processes of LLMs.

In this paper, we propose Knowledge Graph-based Retrofitting (KGR), a new framework that incorporates LLMs with KGs to mitigate factual hallucination during the entire reasoning process of LLMs. Instead of retrieving factual information from KGs using original queries, the main idea behind KGR is to autonomously retrofit the initial draft responses of LLMs based on the factual knowledge stored in KGs. However, achieving the above process is challenging because draft responses generated by large language models typically contain a mixture of various information about the reasoning process, making the extraction, verification, and revision of relevant knowledge in it very challenging. Therefore, the key to integrating Knowledge Graphs into the reasoning process of large models to mitigate factual hallucinations lies in efficiently extracting the information requiring validation from draft responses, querying and selecting relevant knowledge from the knowledge graphs, and using this knowledge to verify and refine draft responses. 

To this end, KGR presents a LLMs-based framework to autonomously extract, validate and refine factual statements within the initial draft responses without any manual efforts. Specifically, given an input query and a draft response generated by the LLM that entails the reasoning process of how LLM resolves this problem, KGR will request a LLM to extract factual claims in the reasoning process that require verifying by KGs. As shown in Figure~\ref{fig:flow}, given the draft response ``\textit{Frédéric Chopin's father is Nicolas Chopin, he was born on June 17, 1771.}'', the claim extraction step will generate factual claims in it like ``\textit{Frédéric Chopin's father is Nicolas Chopin}'' and ``\textit{Nicolas Chopin was born on June 17, 1771}''. Then, KGR will identify critical entities in the extracted claims, retrieve relevant factual triples from knowledge graph about the entities, and use a LLM-based fact selector to identify fact triples relevant to the draft response. Subsequently, the retrieved factual knowledge is utilized to compare with the previously extracted factual claims from the draft to verify their correctness. Finally, LLMs are asked to retrofit the draft in accordance with the outcomes of factual verification. This process can be repeated multiple times to ensure that all facts in the generated answers align with the knowledge stored within the knowledge graph.
In this way, our method can not only verify the fact in query and response but also the facts used during reasoning. Furthermore, because all phases in the procedure can be automatically executed using a large language model, our method doesn't need any external components and therefore is easy to implement.

We conduct experiments with three representative LLMs on three standard factual QA benchmarks with different levels of reasoning difficulty, including Simple Question~\cite{bordes2015large}, Mintaka~\cite{sen2022mintaka} for complex reasoning, and HotpotQA\cite{yang2018hotpotqa} for open domain, multi-hop reasoning. Experiments show that KGR can significantly improve the performance of LLMs on factual QA benchmarks especially when involving complex reasoning processes, which demonstrates the necessity and effectiveness of KGR in mitigating hallucination.

In summary, the contributions are as follows:
\begin{itemize}
    \item We propose a new framework that incorporates LLMs with KGs to mitigate factual hallucination by effectively extracting, verifying, and refining factual knowledge in the entire reasoning process of LLMs.
    \item We present an implementation of the above-mentioned procedure by executing all the above-mentioned steps using LLMs without introducing any additional efforts.
    \item Experiments on 3 datasets and 3 different LLMs confirm that KGR can significantly mitigate the hallucination and enhance the reliability of LLMs.
\end{itemize}

\section{Related Work}
\paragraph{Hallucination}
Hallucination in Large Language Models has been a prominent research focus within the NLP community~\cite{ji2023survey}. Automated large-scale data collection processes are prone to collecting erroneous information, which can significantly impact the quality of the generated outputs~\cite{gunasekar2023textbooks}. Additionally, excessive repetition of certain data during training can introduce memory biases, further exacerbating the hallucination issue~\cite{lee-etal-2022-deduplicating,biderman2023pythia}. Imperfections in the encoder backbone and variations in decoding strategies also play a role in determining the extent of hallucination in LLMs outputs~\cite{tian2019sticking}. Recent studies have emphasized the importance of model output confidence as an indicator of potential hallucination occurrences~\cite{manakul2023selfcheckgpt}.

\paragraph{Retrieval Augmentation}

To address hallucination issues in LLMs, two main categories of retrieval augmentation methods have been proposed, which can be concluded as ``retrieve before generation`` and ``retrieve after generation``.
The retrieve before generation mainly focuses on leveraging information retrieval (IR) to provide additional information to LLMs about the query. Along this line, UniWeb~\cite{li2023web} introduces an adaptive method for determining the optimal quantity of referenced web text,
Chameleon\cite{lu2023chameleon} leverages an assortment of tools including search engines, to bolster the reasoning capabilities of LLMs, WebGLM \cite{liu2023webglm} augments LLMs with web search and retrieval capabilities. 
One major limitation of these approaches is the retrieved text is question-related, thus cannot guarantee the correctness of the question-unrelated portions in the generations. 
The retrieve after generation like RARR~\cite{gao2023rarr}, PURR~\cite{chen2023purr}, and CRITIC~\cite{gou2023critic} automatically edit model generations using evidence from the web. Our method leverages KGs as knowledge base to retrofit the model-generated response while reducing hallucination risk.

\paragraph{KG-Enhanced LLM}
The Knowledge Graph is regarded as a dependable source of information and is consequently frequently employed to enhance model generations. Traditional approaches involve knowledge representations during the training phase, which often necessitates dedicated model architecture and model-specific training \cite{zhang-etal-2019-ernie, zhang2022dkplm}. However, this incurs a substantial cost for contemporary LLMs.
Recent years, many researchers propose to inject knowledge while inference. For example, KAPING~\cite{baek2023knowledge}, RHO~\cite{ji2022rho}, KITLM \cite{agarwal2023kitlm}, and StructGPT~\cite{Jiang-StructGPT-2022} try to retrieve knowledge in KG and utilize them as an additional input context for LLMs to enhance their generations. However, these methods only search for question-relevant information, which limits the overall performance. To the best of our knowledge, we're the first to involve knowledge graphs into model response retrofitting.

\section{KGR: Autonomous Knowledge Graph-based Retrofitting}

In this section, we introduce our proposed method KGR, which automatically mitigates factual hallucinations via a chain-of-verification process. As shown in Figure \ref{fig:flow}, given a query and its draft response, KGR retrofits the response by 1) extracting claims from the draft answer that requires verification; 2) detecting entities in the claims that are critical for retrieving facts from knowledge graph; 3) retrieving relevant fact statements from the knowledge graph; 4) verifying the factual correctness of each extracted claim using the returned factual statements from the knowledge graph; 5) retrofitting the previous draft response based on the verification results. All these steps are autonomously executed using the large language model itself without additional manual efforts. And this process can be iterative and repeated multiple times to ensure that all facts in the generated answers align with the factual knowledge stored within the knowledge graph. In the following, we will describe each component in KGR respectively in detail.

\subsection{Claim Extraction}
Given a generated draft response as input, claim extraction will extract all factual claims from previously generated drafts that require validation. The main idea behind claim extraction is that a draft response can frequently contain various factual statements that need to be verified. For the example in Figure~\ref{fig:flow}, the draft response contains at least two factual statements, i.e., ``\textit{Frédéric Chopin's father is Nicolas Chopin}'' and ``\textit{Nicolas Chopin was born on June 17, 1771}''.
Therefore, to make it possible for KG to verify these statements respectively, claim extraction decomposes the draft response to be atomic factual claims.

\begin{figure}[!ht]
    \centering
    \includegraphics[width=0.5\textwidth]{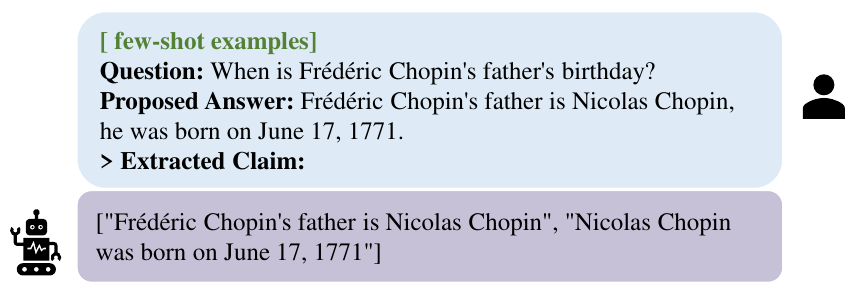}
    \caption{Example for the claim extraction in KGR, which decomposes the proposed answer into two atomic claims.}
    \label{fig:claim}
\end{figure}

Previous work has shown that large language models have strong abilities to extract various kinds of critical information from texts via in-context few-shot learning~\cite{chern2023factool}. Therefore in this paper, we leverage LLM itself to autonomously extract the claims in the generated draft response. As shown in Figure~\ref{fig:claim}, we prompt LLM with a query and response pair, with the anticipation of receiving a list of decomposed factual claims. 

After extracting claims, entity detection identifies mentioned critical entities for knowledge graph retrieval.

\subsection{Entity Detection and Knowledge Graph Retrieval}

Given a list of claims extracted from the draft response, entity detection will detect the critical entities mentioned in the claims. Then, we retrieve the detected entities' local subgraph from the KG and expressed it in the form of triples. The main idea behind entity detection and knowledge graph retrieval is that we need to identify entities in claims so as to retrieve the relevant knowledge in the KG.
Meanwhile, we can ensure recalling relevant triples as much as possible by retrieving the local subgraph in the knowledge graph.
For the example in Figure~\ref{fig:flow}, we identify the entity \textit{Frédéric Chopin} and its entity id \textit{Q1268}, so we can search the identified entity to acquire knowledge relevant to \textit{Claim1} in the KG.

Previous methods~\cite{brank2017annotating,Honnibal_spaCy_Industrial-strength_Natural_2020} rely on supervised fine-tuned models which necessitate training for specific knowledge graphs, resulting in poor generalization for different scenarios. Furthermore, it proves to be a challenging task to discern essential entities that merit fact selection.

\begin{figure}[!h]
    \centering
    \includegraphics[width=0.4\textwidth]{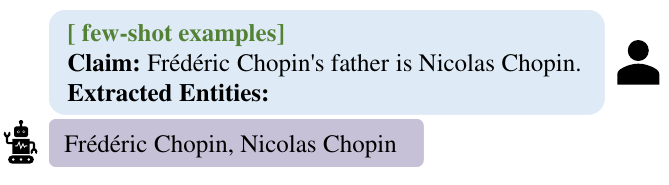}
    \caption{Example for the entity detection in KGR, which only extract the essential entities from the claim.}
    \label{fig:detection}
\end{figure}

In this paper, we prompt LLMs to detect entities. As illustrated in Figure~\ref{fig:detection}, our approach shows powerful generalization ability by capitalizing on the information extraction capabilities of LLMs\cite{li2023evaluating} through the utilization of few-shot prompt.
Based on the few-shot prompts, we can make LLMs understand which entities merit fact selection.

A comparative assessment of performance between the supervised fine-tuned model and LLM is presented in the Appendix.

After detecting the entities, we retrieve the knowledge graph for the local subgraph and send it to fact selection in the form of triples.

\subsection{Fact Selection}
\begin{figure}[!h]
    \centering
    \includegraphics[width=0.4\textwidth]{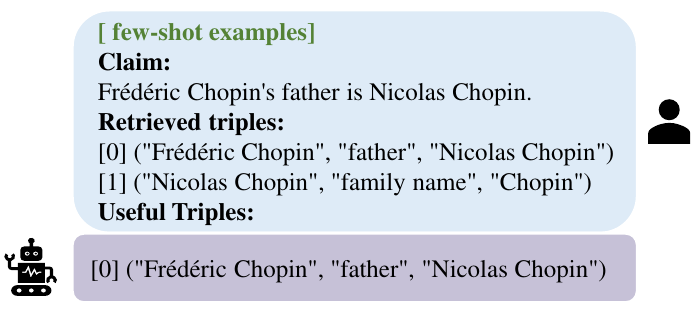}
    \caption{Example for the fact selection in KGR, in which LLMs are prompted to select critical items among all retrieved triples.}
    \label{fig:retrieval}
\end{figure}

Given the retrieved triples based on the detected entities, fact selection will select relevant fact statements among them. 
The main idea behind fact selection is the limited ability of LLMs in long-context modeling ~\cite{liu2023lost} and the constraint on the context window size of LLMs, which make it impractical to select critical triples at one time.
In this paper, we partition retrieved triples into several chunks and leverage LLM itself to extract the critical triples in the retrieved triples respectively, illustrated in Figure \ref{fig:retrieval}. In this way, we can avoid introducing excessive irrelevant knowledge into claim verification.

Previous approaches \cite{cao-etal-2022-kqa} typically relied on text-to-SQL models to formulate query statements for interacting with KG. This method is challenging due to the requirement for substantial annotated training data, which also varies across different KGs. Additionally, it is still challenging to generate compilable and structured SQL, limiting the number of recalled triples.
In contrast, our approach leverages the LLM's information extraction ability, to improve the recall of critical triples.

Once we have selected the critical triples, the claim verification will verify the factual correctness of claims and subsequently offer suggestions.

\subsection{Claim Verification}

\begin{figure}[!h]
    \centering
    \includegraphics[width=0.5\textwidth]{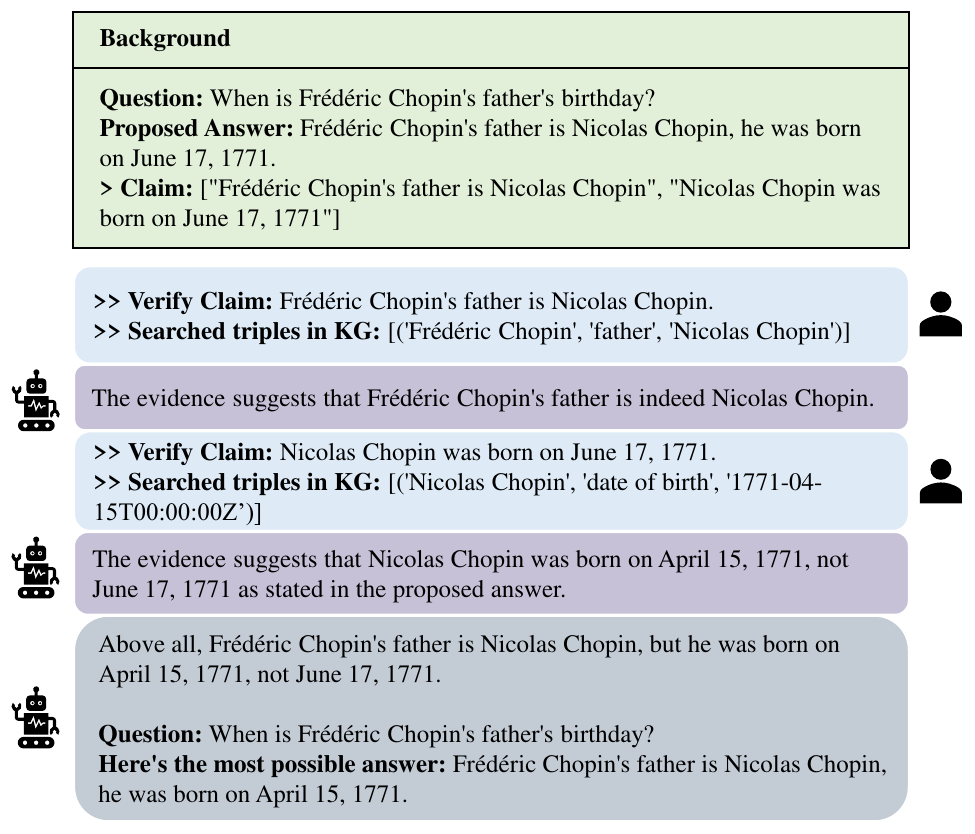}
    \caption{Example for the claim verification and response retrofitting in KGR. The claim verification judges whether the claim aligns with searched triples and gives revision suggestions respectively. The response retrofitting incorporates the revision suggestions from all claims and gives a refined response.}
    \label{fig:verification}
\end{figure}

Given the critical triples selected by the fact selection, we utilize LLM to compare the model-generated claims with the factual information present in the KGs. 
The main idea behind claim verification is to propose a detailed revision suggestion for each claim, as retrofitting solely based on the selected knowledge may not convince LLMs.
As illustrated in Figure~\ref{fig:verification}, we employ LLMs to verify each claim and propose revision suggestions respectively based on the retrieved fact knowledge, so as 
 to boost the execution of the following retrofitting step.

Then, we send the claim verification result to LLM to ask it to retrofit the draft response accordingly.

\subsection{Response Retrofitting}

Given the verification of all claims, the response retrofitting step retrofits the generated draft response in accordance with the verification suggestions.

In this paper, we capitalize on the capabilities of LLMs for the purpose of retrofitting. 
This approach involves employing LLMs with a few-shot prompt, a strategy that has exhibited efficacy in prior researches~\cite{gou2023critic,zheng2023can}.
As illustrated in Figure~\ref{fig:verification}, we merge the entire KGR process into a singular prompt. This allows LLMs to leverage their in-context learning ability, comprehending the KGR process and enhancing their comprehension of factual retrofitting based on verification suggestions

By following the cycle of  ``Extraction - Detection - Selection - Verification - Retrofitting ", our KGR framework can be iterated multiple times to ensure all facts in the generated answers align with the factual knowledge stored within the knowledge graph. 

\section{Experiments}
We evaluate our KGR framework on three datasets with different levels of reasoning difficulty, including Simple Question~\cite{bordes2015large}, Mintaka~\cite{sen2022mintaka}, and HotpotQA~\cite{yang2018hotpotqa}. We also compare KGR with information retrieval-based approaches and previous question-relevant knowledge graph retrieval approaches.

\begin{table*}[!h]
\centering
\resizebox{\linewidth}{!}{\begin{tabular}{ccccccccc}
\toprule
          & \multicolumn{2}{c}{\textbf{Simple Question}}   && \multicolumn{2}{c}{\textbf{Mintaka}} && \multicolumn{2}{c}{\textbf{HotpotQA}} \\ \cmidrule{2-3}  \cmidrule{5-6}  \cmidrule{8-9}
          & \textit{ChatGPT} & \textit{text-davinci-003} && \textit{ChatGPT} & \textit{text-davinci-003} && \textit{ChatGPT} & \textit{text-davinci-003} \\ \midrule
Vanilla   & 22.0/28.9     & 34.7/45.1           && 42.9/56.1 & 36.7/44.8        && 18.4/31.6  & 22.4/34.6        \\ \midrule
CoT       & 10.0/11.8     & 38.0/46.7           && 53.1/59.3 & 46.3/57.9        && 24.5/34.3  & 29.2/40.5        \\
CRITIC    & 12.0/14.3     & 38.0/46.7           && 51.0/58.6 & 44.4/54.0        && 30.6/\textbf{41.7}  & 27.1/38.9        \\
QKR  & 54.0/60.2     & \textbf{56.0}/\textbf{62.0}           && 48.0/54.6 & 44.0/51.7        && 28.0/38.1  & 22.0/31.9        \\
KGR (ours)      & \textbf{58.0}/\textbf{60.7}     & 54.0/57.6           && \textbf{53.1}/\textbf{60.8} & \textbf{52.0}/\textbf{60.2}        && \textbf{32.7}/39.2  & \textbf{34.0}/\textbf{47.2}        \\ \bottomrule
\end{tabular}}
\caption{\textbf{Results on three datasets using ChatGPT and text-davinci-003.} We implement CoT using the prompt provided by CRITIC. QKR uses the same entity detection and fact selection method as KGR. We report both EM and F1 scores in the table.}
\label{table:main-result}
\end{table*}

\subsection{Experiment Settings}

\paragraph{Dataset and Evaluation} We conduct experiments on three representative factual QA benchmarks, including:

\begin{itemize}
    \item \textbf{Simple Question~\cite{bordes2015large}} is a simple QA dataset that contains 100k questions constructed from Freebase knowledge graph. All questions in the dataset are simple, require no deep reasoning procedure, and can be easily answered as long as the correct evidence is retrieved. Therefore, we can evaluate the ability to retrieve relevant evidence in KG based on Simple Question.
    \item \textbf{Mintaka~\cite{sen2022mintaka}} is a complex, natural and multilingual dataset, composing 20k questions collected in 8 different languages. We only use English test sets. In this setting, we focus on the ability to logically refine and revise its answers based on the evidence gathered. 
    \item \textbf{HotpotQA~\cite{yang2018hotpotqa}} is a Wikipedia-based~\footnote{\url{https://wikipedia.org}} dataset with 113k questions that requires finding and reasoning over multiple supporting documents to answer, which are diverse and not constrained to any pre-existing knowledge bases or knowledge schemas. Therefore, we evaluate the KGR framework's robustness in handling generalized scenarios, requiring LLMs to answer involving the incorporation of both parametric knowledge and Knowledge Graph-based information.
    
\end{itemize}
We reported the results in terms of EM and F1 scores respectively on 50 samples from the validation set of each dataset. By comparing performance across these three datasets,
we can evaluate how well different methods mitigate factual hallucinations and handle complex tasks.

\begin{table}[]
\centering
\begin{tabular}{@{}cccc@{}}
\toprule
    & \textbf{Simple Question} & \textbf{Mintaka} & \textbf{HotpotQA}  \\ \midrule
CoT & 14.0/21.9       & 26.0/28.3 & 12.0/19.6 \\
QKR & 40.0/44.0       & \textbf{26.5}/32.4 & \textbf{12.2}/17.6\\
KGR(ours) & \textbf{46.0}/\textbf{46.9}       & \textbf{26.5}/\textbf{34.0} & 10.2/\textbf{20.6} \\ \bottomrule
\end{tabular}
\caption{\textbf{Results on three datasets using Vicuna 13B.} We report both EM and F1 scores in the table.}
\label{table:vicuna}
\end{table}

\paragraph{LLMs and KG Implementation}
We evaluate the effectiveness of KGR on both close-source and open-source large language models. For close-source models, we evaluate on text-davinci-003 and ChatGPT (gpt-3.5-turbo-0301) to see whether alignment will have an impact on KGR. For the open-source model, we evaluate KGR on Vicuna 13B, a representative aligned open-source model, to see whether KGR can work well on compact size LMs.
We choose Wikidata\footnote{\url{https://www.wikidata.org/}}\cite{vrandevcic2014wikidata} as our knowledge base, which encompasses structured data from various sources such as Wikipedia, Wikimedia Commons\cite{wikimedia}, and other wikis associated with the Wikimedia movement\cite{wikidata-movement}.

\subsection{Baseline}
We compared our KGR framework with the following methods, including:
\begin{itemize}
\item {Vanilla, which adopts a straightforward approach to prompt the model to generate answers for the given question. }

\item {Chain of Thought (CoT)~\cite{wei2022chain}, which aims to generate more reliable answers by prompting LLMs to generate more comprehensive and detailed explanations for the generated answers.}

\item { Self-Correcting with Tool-Interactive Critiquing (CR\-/ITIC)} ~\cite{gou2023critic}, which revises the answer based on text from the web. Since CRITIC did not release their web crawling method, in this experiment we adopt the crawling pipeline provided in RARR~\cite{gao2023rarr} via Bing Search\footnote{\url{https://www.bing.com/}.}

\item {Question-relevant Knowledge Retrieval method (QKR), which prompts LLMs with the question-relevant retrieved facts in the knowledge graph to generate answers. In this setting, we aim to demonstrate the superior effectiveness of our response-relevant retrofitting method over the question-relevant knowledge graph augmentation approach. The QKR method, inspired by KAPING~\cite{baek2023knowledge}, leverages extracted facts from KGs as prompts to enhance response correctness. In our implementation, we replace the fact extracts process with our entity detection and fact selection to strictly compare the difference between the response-relevant and query-relevant methods.}

\end{itemize}

\begin{table*}[!h]
\centering
\resizebox{\linewidth}{!}{\begin{tabular}{@{}p{1cm}p{3.5cm}p{6.3cm}p{5.5cm}p{4.5cm}@{}}
\toprule
\multicolumn{5}{p{20cm}}{\textbf{Question}: The city where Alex Shevelev died is the capital of what region?}   
                                \\
\multicolumn{5}{l}{\textbf{Answer}: the Lazio region} 
                                \\
\multicolumn{5}{p{22cm}}{\textbf{Initial prediction}: Let's think step by step. \colorbox{first_one}{Alex Shevelev} died in \colorbox{red_one}{Moscow}, Russia. And it is the capital of the Central Federal District. So the answer is: Central Federal District.}                                                            \\ \toprule
                           & \makecell{Claims}  & \makecell{Fact Knowledge} & \makecell{Verification} & \makecell{Retrofitted Response}               \\ \midrule
\multirow{5}*{\makecell{KGR\\round1}}     &\colorbox{first_one}{Alex Shevelev} died in Moscow, Russia. & (\colorbox{first_one}{Alex Shevelev}, place of death, Rome) & The evidence suggests Alex Shevelev died in \colorbox{red_one}{Rome}, not Moscow, Russia.    & \multirow{5}{=}{Let's think step by step. Alex Shevelev died in \colorbox{red_one}{Rome, Italy}. And it is the capital of the Central Federal District. So the answer is: Central Federal District.}    \\ \addlinespace  \cdashline{2-4}[5pt/2pt] \addlinespace 
                            &\colorbox{second_one}{Moscow} is the capital of the Central Federal District.& (Central Federal District, capital, \colorbox{second_one}{Moscow})
                            &The evidence suggests Moscow is the capital of the Central Federal District.        &                              \\  \midrule
\multirow{5}*{\makecell{KGR\\round2}} &\colorbox{first_one}{Alex Shevelev} died in \colorbox{third_one}{Rome}, Italy.& (\colorbox{first_one}{Alex Shevelev}, place of death, \colorbox{third_one}{Rome})       & The evidence suggests Alex Shevelev died in Rome.       & \multirow{5}{=}{Let's think step by step. Alex Shevelev died in Rome, Italy. And it is the capital of \colorbox{red_one}{Lazio}. So the answer is: Lazio}. \\ \addlinespace   \cdashline{2-4}[5pt/2pt] \addlinespace 
                            &\colorbox{third_one}{Rome} is the capital of the Central Federal District.& (\colorbox{third_one}{Rome}, capital of, Lazio)           &  The evidence suggests Rome is the capital of \colorbox{red_one}{Lazio} not the Central Federal District.      &                              \\ \bottomrule
\end{tabular}}
\caption{Gnerated examples in HotpotQA by ChatGPT. We show the multi-turn retrofitting process of KGR. The  \colorbox{red_one}{red color}  refers to retrofitted factual statements. The other green colors refer to critical entities in claims. }
\label{table:case}
\end{table*}

\subsection{Overall Results}

As shown in Table~\ref{table:main-result}, our method demonstrates significant superiority over other methods across various conditions.

\begin{enumerate}[1)]
    \item \textbf{Our framework can mitigate large language model hallucination via Knowledge Graph-based Retrofitting and achieve significant improvements on 3 datasets.} Compared with the CoT and CRITIC baseline, our KGR framework gains improvements on all three datasets. This indicates that our KG-based approach is more effective due to its reliance on a reliable knowledge base, whereas IR-based methods like CRITIC might introduce noise from external. Additionally, we observed that the CoT method performed worse than the vanilla approach in ChatGPT. This could be attributed to the CoT method's tendency to ask for more information, which is amplified in ChatGPT due to Reinforcement Learning from Human Feedback~\cite{ouyang2022training}.
    \item \textbf{By verifying the facts used during reasoning via chain-of-verification, our method can achieve significant performance improvement in complex reasoning tasks in Mintaka and HotpotQA datasets.} As shown in Table~\ref{table:main-result}, compared to the QKR method, our KGR framework achieves F1 improvement for at least 6.2 and 1.1 on Mintaka and HotpotQA. Both of them pose complex reasoning question-answering challenges, and the success of our method with chain-of-verification on these datasets demonstrates its capability to handle complex questions effectively. It is worth noting that the text-davinci-003 outperformed QKR in Simple Question. We attribute this to the fact that Simple Question consists of straightforward, one-hop questions, which makes the question-relevant method more effective.
    \item \textbf{By automatically generating and executing chain-of-verification via LLMs, our KGR approach exhibits remarkable generalization capabilities in different datasets and is robust on open-domain setting.} In HotpotQA, KGR performs well compared to the CoT and CRITIC methods. The HotpotQA presents an open-domain QA scenario where finding related triples in the KG can be challenging. Despite this difficulty, our method displayed the ability to effectively utilize the searched triples and effectively leverage parametric knowledge even when no evidence was returned.
    \item \textbf{Our framework can work well on compact size LMs, aligned LLMs, and misaligned LLMs, showing the generalizability of KGR.} We compare KGR with the strong baselines CoT and QKR on Simple Question, Mintaka, and HotpotQA using Vicuna 13B. The result is shown in Table~\ref{table:vicuna}. We can find that the KGR framework outperforms both CoT and QKR, demonstrating the generalizability of our framework even leveraging a compact size LM. Moreover, the significant improvement with  ChatGPT and text-davinci-003 shows the generalizability of both aligned LLMs and misaligned LLMs.
\end{enumerate}

In summary, our method consistently outperforms other methods across various conditions and exhibits strong generalization ability. The results suggest that our KGR framework is more reliable and effective, especially in handling complex factual reasoning tasks. Furthermore, it showcases the robustness of our method in open-domain QA settings, where knowledge retrieval may be more challenging.

\subsection{Case Study}
We present a multi-round retrofitting process of a multi-hop case which needs to be retrofitted iteratively in Table~\ref{table:case}. 
In this case, the model-generated response shows a factual error in the initial reasoning step.  It erroneously states that \textit{Alex Shevelev died in Moscow, Russia}, whereas he actually passed away in Rome, Italy. After retrofitting this mistake, we encounter another factual error, which asserts that \textit{Rome is the capital of the Central Federal District}. So, we need to retrofit it again based on the retrofitted response in the first iteration.

From this case, we show KGR's intermediate results, including atomic claim, critical triples, detailed verification, and iterative retrofitting. All these show the effectiveness of KGR, especially on reasoning with multi-hop complex tasks, verifying the feasibility of multi-turn retrofit to ensure that all facts in the generated answers align with the factual knowledge stored in the knowledge graph.

\subsection{Error Analysis}
\begin{figure}
    \centering
    \includegraphics[width=0.35\textwidth]{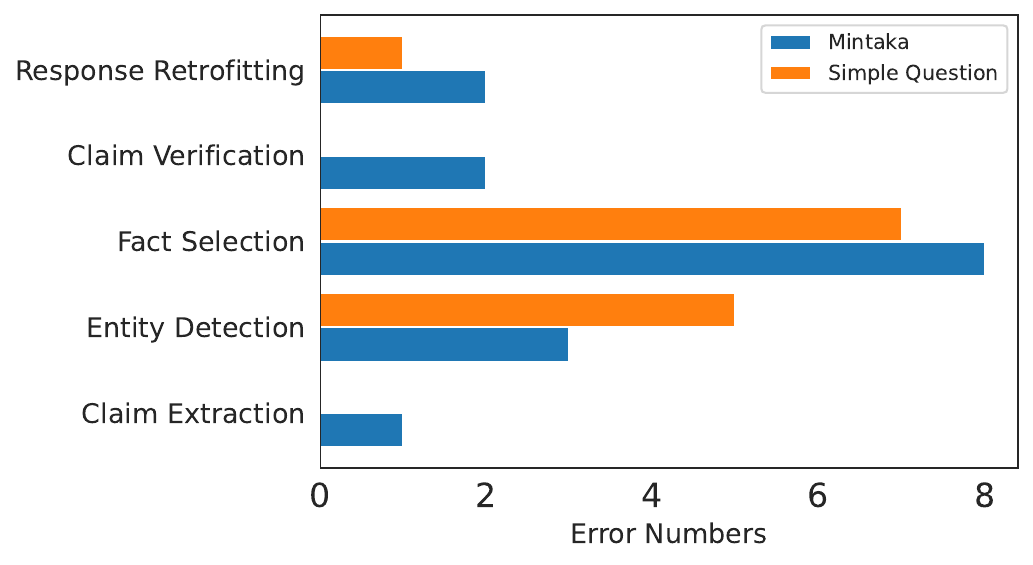}
    \caption{\textbf{Distribution of error case numbers across KGR stages}: Analysis conducted on a sample of 50 instances from the Mintaka dataset and 50 instances from the Simple Question dataset reveals the occurrence of error numbers across various stages of the KGR process.}
    \label{fig:error-analysis}
\end{figure}
In order to gain a comprehensive understanding of the KGR approach, we conducted an exhaustive analysis of incorrect cases based on the Mintaka and Simple Question datasets. 
After carefully examining the errors, we identified which component causes revision failures.
The outcomes of this analysis are visualized in Figure ~\ref{fig:error-analysis}.

On closer inspection, it becomes apparent that inaccuracies within the KGR are primarily caused by entity detection and fact selection. 
Conversely, claim extraction, claim verification, and response retrofitting demonstrate higher reliability. All these findings highlight the significance of refining entity detection and fact selection for further improvements. 

On the other hand, our analysis explored the error reason in different stages. The claim extraction often fails to express the central claim adequately, sometimes due to excessive use of pronouns that confuse the model's comprehension. 
Entity detection  has issues with entity extraction granularity. It captures too many common entities like "films" or "apple", leading to excessive and useless triples for the claim verification.
The fact selection has difficulty extracting the critical triples between multiple triples that contain noise. 
For the claim verification and response retrofitting, the focus shifts to the model's ability to adhere to the cues provided by the few-shot prompts. Effectively discerning and subsequently rectifying answers within this framework presents a central challenge.
The process of fact selection encounters challenges in extracting essential triples from a collection of triples that include irrelevant information or noise.

For a deeper insight into the effectiveness of fact selection, we conduct experiments on it. The effectiveness of entity detection will be shown in Appendix.

\begin{figure}[!h]
\centering 
\subfigure[Results for fact selection conducted on a sample of 50 instances from Simple Question with different chunk sizes. ] {
\includegraphics[width=3.80cm]{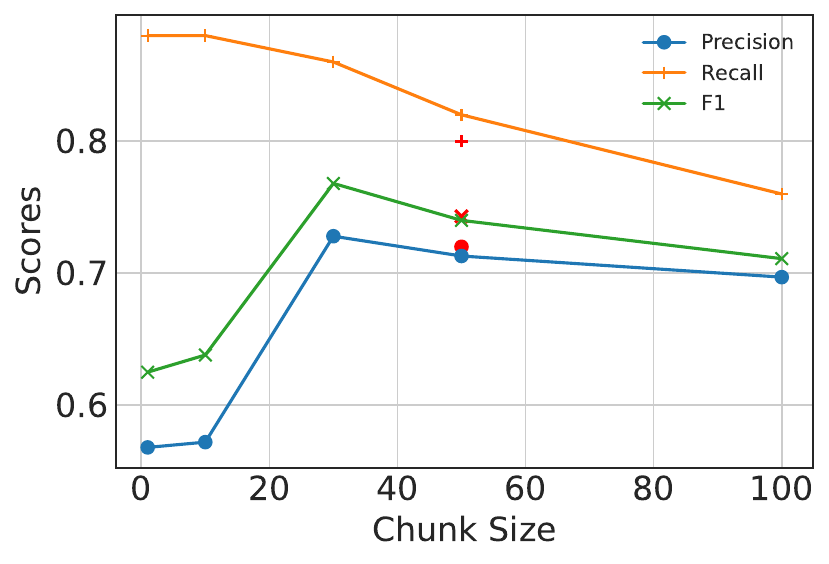}
\label{fig:chunk-size} 
}
\hfill 
\subfigure[Results for fact selection conducted on a sample of 50 instances from Simple Question with different retrieved triple numbers.] {
\includegraphics[width=3.80cm]{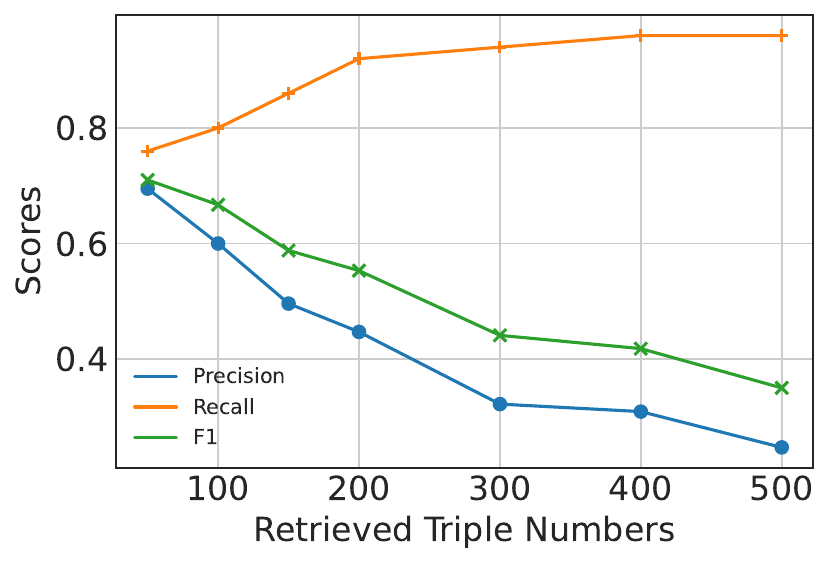} 
\label{fig:triple} }
\caption{Impact of chunk size and numbers of retrieved triples on the efficacy of the fact selection.}

\end{figure}

\paragraph{Impact of chunk size\&numbers of retrieved triples.}
As discussed above, considering the limitation of maximum input length for LLMs, we partition the retrieved triples into chunks for fact selection. However, it is worth noting that in-context learning might be influenced by example order~\cite{lu-etal-2022-fantastically} and potentially following the last answer presented~\cite{zhao2021calibrate}. 
Under these motivations, we conduct experiments on the Simple Question using ChatGPT.
Specifically, we evaluate the effectiveness of fact selection when retrieved triples are in random order, referring red point in Figure~\ref{fig:chunk-size}.
 These experiments help us understand fact selection behavior under various hyperparameters, optimize chunk size, and refine triple retrieval strategies for improved efficiency.

As shown in Figure~\ref{fig:chunk-size}, the chunk size has minimal impact on triple selection capability, except for a chunk size of 100, which may cause worse long-distance dependency modeling. However, reducing the chunk size leads to lower precision and higher recall scores. This indicates that a smaller chunk size increases the chance of selecting both critical and irrelevant triples. 
Additionally, we observe that prompting LLMs with triples in random orders doesn't significantly affect triple selection.

As shown in Figure~\ref{fig:triple}, increasing the number of retrieved triples has a gradual positive impact on recall but significantly reduces precision.
More retrieved triples may boost recall for critical knowledge and introduces numerous irrelevant triples, potentially compromising the effectiveness of the claim verification and negating the benefits of the fact selection. 

All in all, the experiments focusing on the impact of chunk size and numbers of retrieved triples show that the core difficulty of retrieving fact knowledge based on LLMs is the tradeoff between precision and recall. This observation points to future research on fact selection based on LLMs.

\section{Conclusion}
In this paper, we propose a knowledge graph-based retrofitting framework that effectively mitigates factual hallucination during the reasoning process of LLMs based on the factual knowledge stored in KGs. 

Experiment results show that KGR can significantly improve the performance of LLMs on factual QA benchmarks especially when involving complex reasoning, which demonstrates the necessity and effectiveness of KGR in mitigating hallucination and enhancing the reliability of LLMs. As for future work, we plan to improve the effectiveness in each step of our KGR framework.

\appendix
\bibliography{aaai24}

\section{Appendix}
\subsection{Impact of Multi-Turn Retrofitting}
From the above case, it can be observed that one potential drawback of employing the search engine-based retrofitting approach is the potential inaccuracy in the retrieved information. This could consequently lead to an erroneous revision of factually correct claims. This problem becomes more pronounced during multi-turn retrofitting.

\begin{figure}[!h]
    \centering
    \includegraphics[width=0.4\textwidth]{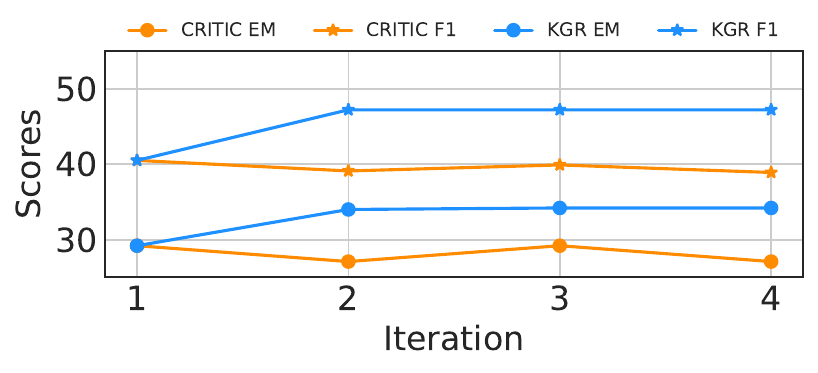}
    \caption{Behavior of KGR and CRITIC under multi-turn retrofitting}
    \label{fig:iteration}
\end{figure}

To investigate the impact of multi-turn retrofitting, we performed multiple rounds of iterative revisions on identical questions from the HotpotQA dataset using text-davinci-003 on both CRITIC and KGR. The outcomes are depicted in Figure~\ref{fig:iteration}. Our observations are as follows: 
(1) Most of the factual errors can be retrofitted in the first turn in our KGR framework.
(2) Within our KGR framework, the effectiveness of the retrofitting process tends to remain consistent owing to the reliability of the underlying knowledge graph. However, the CRITIC method shows a tendency to exhibit poorer performance as the number of iteration turns increases.
By checking the error cases, we note that due to the inherent randomness of information retrieval, the evidence recalled by the search engine for claim verification can vary between rounds. This divergence leads to varying outcomes, introducing disparities in perspectives and factual accuracy, which subsequently leads to fluctuations in the final refined responses.
In contrast, the utilization of a knowledge graph as the foundational knowledge source offers reliable and accurate knowledge information.

\subsection{Effectiveness with Entity Detection.}

As outlined in the Method Section, entity detection is responsible for identifying entities that make sense in claim verification. In Table~\ref{table:entity-linking}, we present a comparison between our method and current widely-used approaches, Wikifier~\cite{brank2017annotating} and SpaCy~\cite{Honnibal_spaCy_Industrial-strength_Natural_2020}, to highlight the advantages of our proposed method.
Wikifier employs a pagerank-based technique to identify a coherent set of relevant concepts, designed for Wikipedia entities.
SpaCy offers a suite of functionalities such as tagging, parsing, named entity recognition, and text classification, leveraging recent advancements in both speed and neural network models. 
The result shows that our LLM-based method delivers superior performance on the Mintaka dataset, validating our decision to leverage the extensive capabilities of LLMs. However, despite the remarkable achievements of LLMs in comparison to alternative techniques, their effectiveness in extracting entities for claim verification remains insufficient. This presents an opportunity for future research in this area.

\begin{table}[!htb]
\centering
\resizebox{\linewidth}{!}{\begin{tabular}{@{}cccccc@{}}
\toprule
   & Wikifier & LLM  & SpaCy & Mix of Wiki+LLM & Mix of all \\ \midrule
Precision  & 0.36     & 0.45 & 0.32  & 0.40         & 0.30      \\
Recall  & 0.34     & 0.6  & 0.56  & 0.68         & 0.74      \\
F1 & 0.33     & 0.48 & 0.39  & 0.47         & 0.40      \\ \bottomrule
\end{tabular}}
\caption{Effect of entity detection using Various Methods. Mix means a hybrid approach among Wikifier, LLM, and SpaCy.}
\label{table:entity-linking}
\end{table}

\end{document}